\pdfoutput=1

\documentclass[conference,letterpaper]{IEEEtran}
\usepackage{algorithm}
\usepackage{algorithmic}
\usepackage[utf8]{inputenc} 
\usepackage[T1]{fontenc}
\usepackage{url}
\usepackage{ifthen}
\usepackage{cite}
\usepackage[cmex10]{amsmath}

\usepackage{amssymb,amsthm}
\usepackage{mathtools} 
\usepackage{mleftright} 
\usepackage{bm,bbm} 

\pdfoutput=1

\def\mc{\mathcal}
\def\mbf{\mathbf}

\usepackage[normalem]{ulem}
\newcommand\redout{\bgroup\markoverwith{\textcolor{red}{\rule[.5ex]{2pt}{0.4pt}}}\ULon}

\DeclareMathOperator*{\argmin}{\mathrm{argmin}}

\DeclareFontFamily{U}{matha}{\hyphenchar\font45}
\DeclareFontShape{U}{matha}{m}{n}{
      <5> <6> <7> <8> <9> <10> gen * matha
      <10.95> matha10 <12> <14.4> <17.28> <20.74> <24.88> matha12
      }{}
\DeclareSymbolFont{matha}{U}{matha}{m}{n}
\DeclareFontSubstitution{U}{matha}{m}{n}

\DeclareFontFamily{U}{mathx}{\hyphenchar\font45}
\DeclareFontShape{U}{mathx}{m}{n}{
      <5> <6> <7> <8> <9> <10>
      <10.95> <12> <14.4> <17.28> <20.74> <24.88>
      mathx10
      }{}
\DeclareSymbolFont{mathx}{U}{mathx}{m}{n}
\DeclareFontSubstitution{U}{mathx}{m}{n}

\newcommand{\reject}{r}
\newcommand{\mobile}{m}
\newcommand{\edge}{e}
\newcommand{\localy}{\hat{y}_{\textrm{local}}}
\newcommand{\edgey}{\hat{y}_{\textrm{server}}}
\newcommand{\rejectB}{r^{B}}

\newcommand{\edgeB}{e^{B}}

\usepackage{tabularx}

\usepackage{multirow}
\usepackage{multicol}

\newtheorem{theorem}{Theorem}

\begin{document}
\title{Learning To Help: Training Models to Assist Legacy Devices}

\author{%
  \IEEEauthorblockN{Yu~Wu}
  \IEEEauthorblockA{Department of Electrical and Computer Engineering\\
                    Rutgers, The State University of New Jersey\\
                    Piscataway, NJ, USA\\
                    Email: yu.wu@rutgers.edu}
  \and
  \IEEEauthorblockN{Anand D.~Sarwate}
  \IEEEauthorblockA{Department of Electrical and Computer Engineering\\
                    Rutgers, The State University of New Jersey\\
                    Piscataway, NJ, USA\\
                    Email: anand.sarwate@rutgers.edu}
}

\maketitle

\pdfoutput=1

\begin{abstract}
Machine learning models implemented in hardware on physical devices may be deployed for a long time. The computational abilities of the device may be limited and become outdated with respect to newer improvements. Because of the size of ML models, offloading some computation (e.g. to an edge cloud) can help such legacy devices. We cast this problem in the framework of learning with abstention (LWA) in which the expert (edge) must be trained to assist the client (device). Prior work on LWA trains the client assuming the edge is either an oracle or a human expert. 
In this work, we formalize the reverse problem of training the expert for a fixed (legacy) client. As in LWA, the client uses a rejection rule to decide when to offload inference to the expert (at a cost). 
We find the Bayes-optimal rule, prove a generalization bound, and find a consistent surrogate loss function. Empirical results show that our framework outperforms confidence-based rejection rules.

\end{abstract}
\pdfoutput=1

\section{Introduction}
\label{sec:intro}
The emerging paradigm of \emph{mobile edge cloud (MEC)} systems provides a rich source of interesting machine learning problems. In a MEC architecture, \emph{mobile devices} are assisted by cloud computing resources at the \emph{``edge''} of the network. Mobile devices have computation and energy consumption limitations that are not shared by the edge. Inference using modern machine learning (ML) models is computationally intensive, so one proposal is for mobile devices to offload these computations to the edge. This comes at cost because the communication introduces additional latency which is undesirable in real-time applications such as augmented reality (AR) and autonomous driving.

Learning with abstention (LWA) or learning with a reject option is a framework which captures many aspects of this scenario: a local client (the mobile device) can abstain from making a prediction on an instance and send it to an expert (the edge server) instead. Prior works generally either discard the abstained instance (see Figure \ref{fig:primitive}(a) and Section \ref{section:leraning to reject}), or treat the expert as an oracle with perfect accuracy or human expert with certain accuracy for specific instances and incorporate a cost for abstention into a loss function for training the local model (see Figure \ref{fig:primitive}(b) and Section \ref{section: leraning to defer}). The goal is for the local client to \emph{learn to ask for help}.

The MEC scenario presents two departures from the standard LWA setup. First, in applications involving large-scale deployment of physical devices may not be possible to retrain the local models. This is particularly true if models are encoded into hardware, which is more efficient in terms of speed and power. Second, the edge server in reality will not be a perfect oracle or human expert. These two aspects---local models on ``legacy devices'' and fallible expert models---suggests a new problem of training an expert/edge server model to \emph{learn how to help} (see Figure \ref{fig:primitive}(c)).

In this paper we study the problem of ``learning to help'' for the ML task-offloading setting. We consider the binary classification setting where there are three decision functions: a classifier $\mobile(x)$ at the local client, a classifier $\edge(x)$ at the edge server, and a rejection rule $\reject(x)$ at the local client which determines whether or not to offload the task to the server. The goal is to "assist" the legacy classifier $\mobile(x)$ by training a rejection rule $\reject(x)$ and a classifier $\edge(x)$ at edge.

The main contributions of this work are:
\begin{itemize}
    \item we identify the general learning to help framework that coordinate the collaborative prediction for ML task-offloading setting when local classifier is fixed;
    \item we formulate mathematical models and derive the Bayes classifiers for models with ability-constrained client;
    \item we propose one convex and differentiable surrogate loss functions that is consistent and the experimental results demonstrate that models trained by this surrogate loss function outperforms confidence-based methods and imply that learning to help framework makes training process more proactive.
\end{itemize}

\pdfoutput=1

\section{Problem Formulation}
\noindent \textbf{Notation.} Let $[n] = \{1,2,\ldots, n\}$ for a positive integer $n$

\begin{figure}[t]
    \centering
     \includegraphics[width=0.42\textwidth]{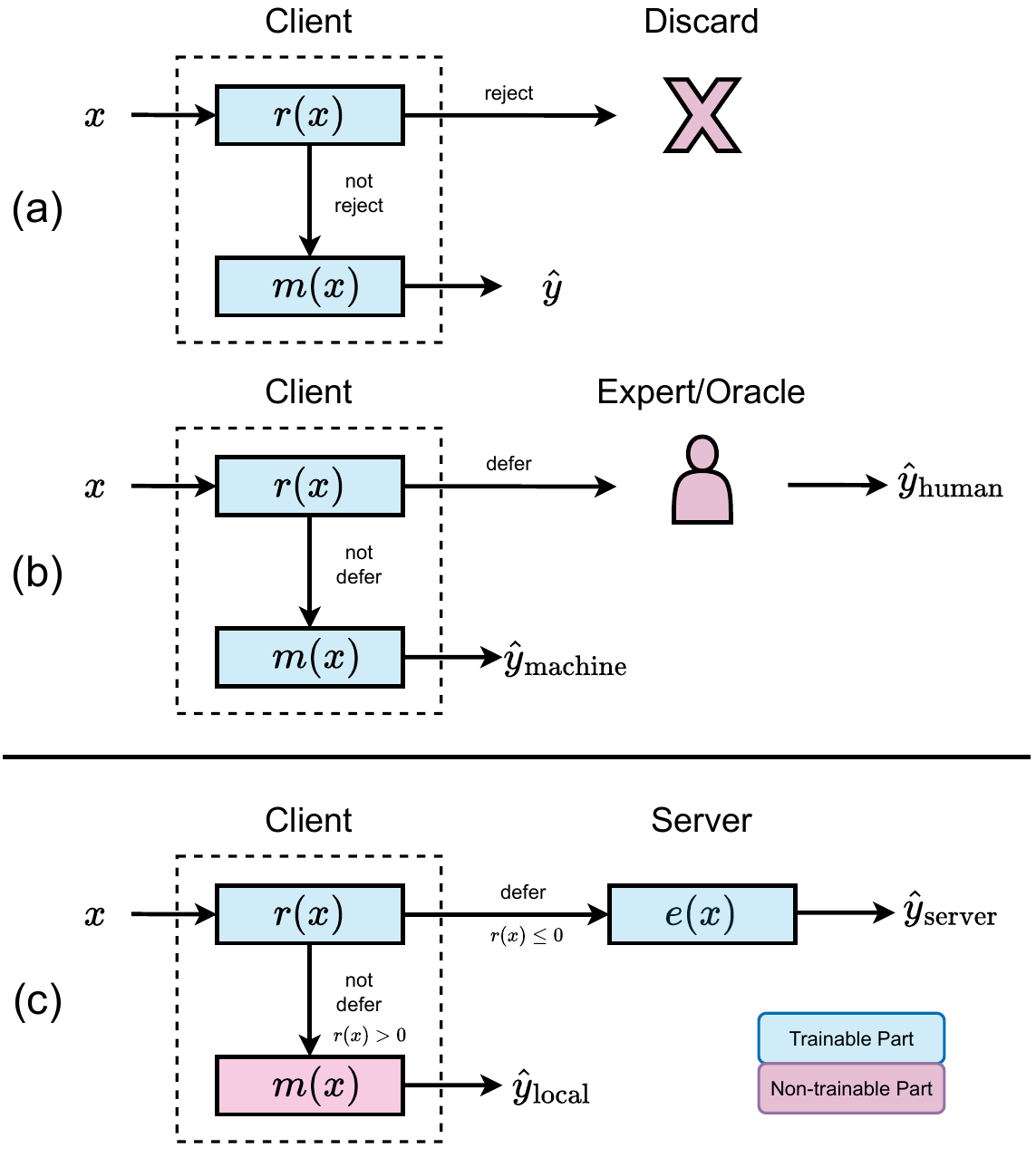}
    \caption{(a) Diagram of the standard learning to reject framework. (b) Diagram of the standard learning to defer framework. (c) Diagram of the learning to help for legacy model framework.}
    \label{fig:primitive}
\end{figure}
We call our problem Learning to Help as shown in Figure \ref{fig:primitive}(c). There are two parties in the system: a \emph{client} and a \emph{server}. The client observes an input $x$ and can either predict a response $\localy$ using a local predictor $\mobile(x)$ or send $x$ to the server, which can predict a response $\edgey$ using a different predictor $\edge(x)$. In our motivating examples, the client may be a device with (relatively) limited computational capabilities whereas the server has more computing resources. This means the server's predictor $\edge(x)$ can be more complex than the client predictor $\mobile(x)$. 
The communication between them is asymmetric (one side can only request limited types of information for the other side) and not free. For this paper we will describe our approach for the simpler case of binary classification, so $\mobile(x)$ and $\edge(x)$ are both classifiers. The input to the system is a feature vector $\mc{X}\subset \mathbb{R}^{k}$ and the output are binary labels in $\mc{Y} = \{-1,+1\}$, without loss of generality. We assume that feature-label pairs in $\mc{X} \times \mc{Y}$ are drawn from an unknown distribution $\mathbf{D}$.

In Figure \ref{fig:primitive}(c) the client's decision can be separated into two functions $r$ and $m$. The function $\reject(x)\colon \mc{X} \to [-1,1]$ is a \emph{rejection rule} (or \emph{rejector}) in the framework of learning with abstention. If $\reject(x) > 0$ then the client will label using the \emph{local rule} $\mobile(x)$ and if $\reject(x) \le 0$ it will sent $x$ to the server to be labeled with $\edge(x)$.  We use the 0-1 loss for the cost on the client. We assume that asking for help is not free: each rejection comes with a cost $c_e$. This cost can be monetary charge (e.g.~for a third-party service running on the server), energy loss for transmission, or latency penalty.
The server's model $\edge(x)$ may also make an error, also measured by the 0-1 loss, but is weighted by a second cost factor $c_1$.
Put together, the generalized 0-1 loss function for this model is
\begin{align}
      L(r, e, x, y) &=\mathbf{1}_{\mobile(x) y \leqslant 0} \mathbf{1}_{\reject(x)>0}+(c_{1}+c_{e})\mathbf{1}_{\edge(x)\neq y} \mathbf{1}_{\reject(x) \leqslant0}\nonumber\\&+c_{e}\mathbf{1}_{\edge(x)= y} \mathbf{1}_{\reject(x) \leqslant0}, \label{c1celLoss}
\end{align}

where $\mathbf{1}_{[\cdot]}$ is the indicator function. The corresponding risk is $R(r,e):=\mathbf{E}_{(X, Y) \sim \mathbf{D}}[L(r, e, x, y)]$. 

The \emph{Bayes-optimal classifiers} are 
\begin{align}
    \rejectB, \edgeB = \argmin_{\reject, \edge} R(\reject, \edge). \label{eq:BayesRule}
\end{align}
The learning problem is to select the functions $\reject(x)$ and $\edge(x)$ based on a training set $\{ (x_i, y_i) : i \in [n]\}$ sampled i.i.d.~from a population distribution $\mbf{D}$. In the following section, we will briefly introduce development phase of the general topic learning to reject, together with the learning to defer. Then we will discuss our main result for Learning to Help in section \ref{section:fix m}.

\pdfoutput=1

\section{Related Work}
\label{section:relteadwork}
\subsection{Learning to Reject}
\label{section:leraning to reject}
Learning with reject option was originally proposed by Chow~\cite{chow1957optimum,chow1970optimum} for a Bayesian pattern recognition system. Decades later,  this problem was revisited by Herbei and Wegkamp, who found upper bounds for a plug-in classification rule\cite{herbei2006classification}. 

Their classification rules relied on the knowledge of data distribution, which is inaccessible in practice. Moreover, the loss functions in earlier works were not differentiable, posing a challenge for optimization-based learning approaches. A series of follow-on works found convex and differentiable surrogate loss functions for this problem~\cite{bartlett2008classification,wegkamp2007lasso,wegkamp2011support,grandvalet2008support}. More recently, a more empirical heuristic using confidence-based rejection models has become popular~\cite{hendrickx2021machine}. The main idea of confidence-based models is to take the output of classifier as a confidence score and the reject option is triggered by the value of this score~\cite{gamelas2015robust,geifman2019selectivenet,jiang2018trust,raghu2019algorithmic}.

The rejector used in above literature is just a function of $\mobile(x)$ instead of $x$. However, those confidence-based methods may fail in special cases and proposed a novel model where separate $\mobile(x)$ and $\reject(x)$ are trained simultaneously~\cite{cortes2016learning,cortes2016boosting}. Moreover, requiring computing $\mobile(x)$ may be a waste of time and energy if the client decides to reject.

\subsection{Learning to Defer}
\label{section: leraning to defer}
One important question was seldom mentioned in previous papers: what will happen after rejection decision? Recent papers linked this topic to Human Computer Interaction (HCI) by consulting human expert when rejected. Most related to our work is a recent paper by Mozannar and Sontag~\cite{mozannar2020consistent}, who firstly formulated the Learning to Defer as a cost-sensitive learning problem with surrogate loss function that can defer to human expert. Follow-on work proposed different surrogate loss functions that are calibrated and consistent to 0-1 loss~\cite{verma2022calibrated,cao2024defense,mozannar2023should} and extended the framework to multiple experts~\cite{keswani2021towards,madras2018predict,hemmer2023learning,mao2024two,verma2023learning}. A related problem is the reverse case where human expert can defer to a (machine) classifier based on mental model~\cite{mozannar2021teaching}.

All the existing papers that related to learning with reject option either consider the abstention as third label or push the data sample to an perfect oracle/human expert. In this work we explore the case where a ``machine'' can ask another ``machine'' for help, and focus on training the helper.

\pdfoutput=1

\section{Learning to Help for a legacy client}
\label{section:fix m}

Consider the case where the local client $\mobile(x)$ is fixed and we must jointly optimize the rejector $\reject(x)$ and server $\edge(    x)$. The following results are based on the assumptions attached in Appendix~\ref{appendix:assumption}.

\subsection{Bayes-optimal decision rules}

While jointly training $\reject(x)$ and $\edge(x)$, the Bayes optimal decision rules for Learning to Help with fixed $\mobile(x)$ are:
\begin{align}
    e^{B}&=\mathrm{sign}\left(\eta(x)-\frac{1}{2}\right) \\
    r^{B}&=c_{1}(\frac{1}{2}-|\eta-\frac{1}{2}|)-P(M\neq Y|X=x)+c_{e},
\end{align}
where $\eta(x)=P(Y=1|X=x)$ is the regression function and $P(M\neq Y|X=x)$ is the error probability for $\mobile(x)$ with given input $x$. The proof is shown in Appendix \ref{proof: bayes rule}. From the Bayes optimal rules above, we know that the the optimal rule for classifier on server is the same as the Bayes classifier for single classifier system. The reject rule is more complicated and is jointly determined by the regression function, cost $c_1$, $c_e$ and error rate of local classifier. If we let $c_1=1$ and $c_e=0$, the reject rule becomes:$r^{B}=P(M=Y|X=x)-1/2-|\eta-1/2|$, which is comparing whether the client or server is more ambiguous (event probabilities close to 1/2). Also, larger $c_1$ and $c_e$ would prevent $\reject(x)$ from asking for because the cost for help is too high while larger $P(M\neq Y|X=x)$ would push $\reject(x)$ to ask for help since $\mobile(x)$ is always inaccurate. Those primary analysis coincide with our regular intuition when making decision. 

\subsection{Generalization Bound}

Learning to Help with fixed $m(x)$ is jointly working with two classifiers $r(x)$ and $e(x)$ located on different components.  We next derive the generalization upper bound in terms of the Radermacher complexity.

\begin{theorem}\label{Bound:natural loss}
Let $\mathcal{R}$ and $\mathcal{E}$ be families of functions that are mapping to $\{-1, +1\}$. Let $e(x)$ be a fixed function that only takes value in $\{-1, +1\}$. We let $R$ denote expected loss of (\ref{c1celLoss}) and $\hat{R}$ denote the empirical expected loss, $n$ is the sample size, $c_{e}$ denotes the cost for asking server classifier and $c_{1}$ denotes the cost when server classifier makes mistake. Then for any $\delta>0$, with probability at least $1-\delta$ over the draw of a set of samples $S$ from $\mathcal{D}$, the following holds for every pair of $(r, e)\in\mathcal{R}\times\mathcal{E}$:
\begin{align}
    R(r, e)&\le \hat{R}(r, e)+(c_{e}+c_{1}+1)\Hat{\mathfrak{R}}_{S}(\mathcal{R})+c_{1}\Hat{\mathfrak{R}}_{S}(\mathcal{E})
    \nonumber \\
    &\hspace{1cm}+3\sqrt{\frac{(c_{1}+c_{e})^{2}\ln{\frac{2}{\delta}}}{2n}},\label{bound:any pair}
\end{align}
where $\Hat{\mathfrak{R}}_{S}(\mathcal{R})$ and $\Hat{\mathfrak{R}}_{S}(\mathcal{E})$ denote the empirical  Radermacher Complexity for $\mathcal{R}$ and $\mathcal{E}$, respectively. Then let $\hat{r}^{*}_{2}, \hat{e}^{*}_{2}$ denote the empirical risk minimizers over the loss function (\ref{c1celLoss}), then for any $\delta>0$, with probability at least $1-\frac{3}{2}\delta$ over the draw of a set of samples $S$ from $\mathcal{D}$, the risk of empirical risk minimizers is upper bounded by the inequality:
\begin{align}
    R(\hat{r}^{*}, \hat{e}^{*})&\le R(r^{B},e^{B})+ (c_{e}+c_{1}+1)\Hat{\mathfrak{R}}_{S}(\mathcal{R})
    \nonumber\\
    &\hspace{1cm}
    +c_{1}\Hat{\mathfrak{R}}_{S}(\mathcal{E})+4\sqrt{\frac{(c_{1}+c_{e})^{2}\ln{\frac{2}{\delta}}}{2n}}.\label{bound: ERM}
\end{align}
\end{theorem}
The proof of Theorem \ref{Bound:natural loss} is attached in Appendix \ref{proof: Rademacher natural}. From (\ref{bound:any pair}) we have that the gap between empirical risk and generalization risk for any pair $(r, e)$ during training is converging with rate $O(\sqrt{1/n})$. While from (\ref{bound: ERM}) we know that the excess risk for empirical minimizer is also converging with rate $O(\sqrt{1/n})$ as well. Both upper bounds depends on the properties of function classes of $r$ and $e$.

\begin{figure*}[ht]
    \centering
     \includegraphics[width=1\textwidth]{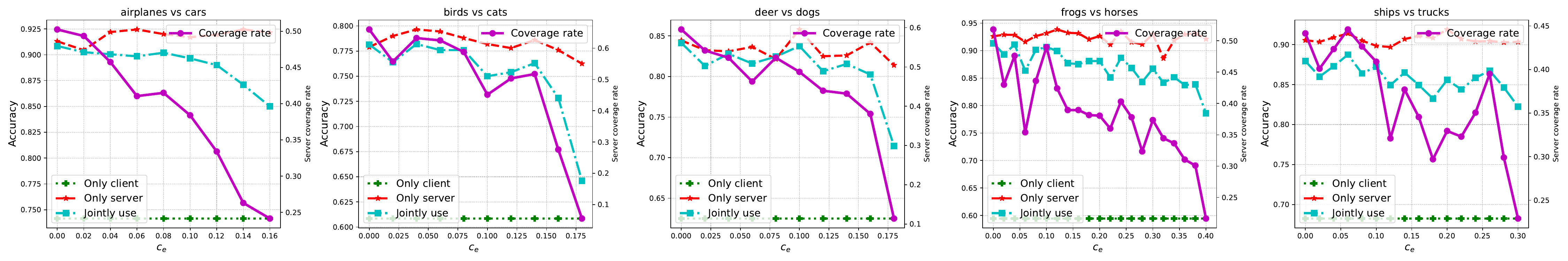}
    \caption{Testing Accuracy and Server Coverage Rate for Different Cost $c_e$. We test the training results for learning to help with fixed local model. Each sub-figure refers to binary classification on two different classes chosen from CIFAR-10.}
    \label{fig:differentcost}
\end{figure*}

\subsection{Surrogate Loss}
The generalized 0-1 loss function with indicators is not convex and differentiable. To make it implementable by optimization techniques, we propose a surrogate loss function:
\begin{align}\label{equ:L2S}
L_{\text{S}}(r,e,x,y)&=c_1 \exp \left(\frac{\beta}{2}(-e y-r)\right)+c_e \exp({-r})
\nonumber\\
&\hspace{1cm} +1_{m(x)y \le 0}\exp \left(\alpha r\right)
\end{align}
where $\alpha$ and $\beta$ are positive real parameters (which can depend on $x$) that are used for calibration. This function is a convex and differentiable ($r$ and $e$) upper bound $L$. In the following theorem, for different situations we assign different calibration parameters, then our surrogate convex loss function is consistent with the generalized 0-1 loss.
\begin{theorem}[Calibration of Surrogate Loss, $L_{\text{S}}$]\label{surrogateLoss3:calibration}
The \(\inf _{(e, r)} \mathbb{E}_{y | x} \mathbb{E}_{m(X) |x, y}[L_{\text{S}} (r,e,x,y) ]\)
is attained at \(\left(e^{*}, r^{*}\right)\) such that \(\operatorname{sign}\left(e^{*}\right)=\operatorname{sign}\left(e^{B}\right)\) and \(\operatorname{sign}\left(r^{*}\right)=\operatorname{sign}\left(r^{B}\right)\) if we let
\begin{itemize}
    \item $0<\alpha\le 1$ and $\beta=2$, for any $x$ that has $P(x)\le c_e$;
    \item  $\alpha >1$ and $\beta=2$, for any $x$ that has $P(x)>c_e+\frac{1}{2}c_1$;
    \item  $\alpha=\frac{c_e+\sqrt{4c_1 P(x)-4c_1 c_e-4 P(x)^2+8c_e P(x)-4c_{e}^{2}}}{P(x)}$ and $\beta=2$, for any $x$ that has $c_e<P(x)\le c_e+\frac{1}{2}c_1$,
\end{itemize}
where $P(x)=P(m(X) \neq Y| X=x)$.
\end{theorem}

The proof of Theorem \ref{surrogateLoss3:calibration} can be found in Appendix \ref{Proof:Calibration}. This theorem provides the guideline to choose calibration parameters during training for each data sample. This result, while theoretically appealing, is impossible to adopt in practice because we do not know $P(x)$ in advance.
To use this result in practice we can estimate $P(x)$ during training. The empirical algorithm for training with our proposed surrogate loss function is attached in Appendix \ref{algo: empirical}. For inference stage, $r(x)$ will directly determine whether to seek help or not and let $e(x)$ or $m(x)$ give the final prediction without the need for $P(x)$ or estimate of $P(x)$ as shown in the inference algorithm \ref{algo: inference}.

\begin{figure*}[htbp]
    \centering
     \includegraphics[width=1\textwidth]{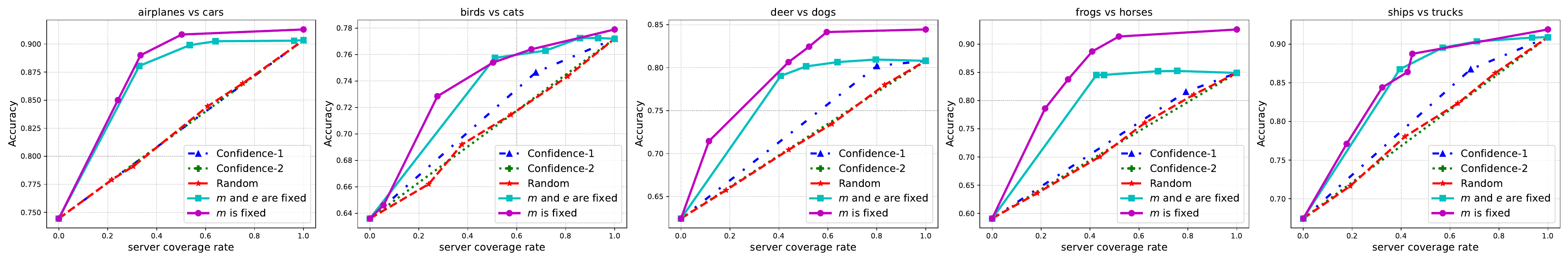}
    \caption{Testing Accuracy Comparison on different Methods over Coverage Rate. We compare accuracy for the surrogate function with other two confidence-based methods as well as randomly reject model. }
    \label{fig:compare}
\end{figure*}

\pdfoutput=1

\section{Experiments}
\label{section:experiment}
In this section, we test our Learning to Help model on public dataset CIFAR-10.  We evaluate how the hyper-parameters influence the training of $\reject(x)$ and $\edge(x)$. We compare the accuracy under the same coverage rate with other confidence-based methods. All experiments are built based on PyTorch.  The CIFAR-10 dataset contains 60000 32x32 color images in 10 different classes. Since our models are designed for binary classification, we divide them into five pairs and test our models on those pairs, respectively.

\textbf{Models Setting.} In most applications of interest, the classifier on client usually should be less ``capable'' than the classifier on the server. The rejector should also be a ``simpler'' function that can make a quick decision without evaluating $\mobile(x)$. We tested a client $\mobile(x)$ set as a neural network with three fully connected layers Also, there is no non-linear activation function for this neural network so it is equivalent as linear regression\cite{jiao2020does}. Server $\edge(x)$ is set as a convolutional neural network with two convolution layers and three fully connected layers. The rejector $\reject(x)$ is a simple neural network with two fully connected layers. For Learning to Help with fixed $\mobile(x)$, we firstly pre-train $\mobile(x)$ with one epoch and then keep it fixed during the jointly training of $\reject(x)$ and $\edge(x)$ for 10 epochs. As for contrast, we also consider the case where both $\mobile(x)$ and $\edge(x)$ are pre-trained for 10 epochs and only train $\reject(x)$. For confidence-based methods, we use the same pre-trained $\mobile(x)$ and $\edge(x)$, and choose to add sigmoid function to the tail of $\mobile(x)$ and let the output work as confidence score.\cite{geifman2019selectivenet} designs distance metric $\mobile(x)(1-\mobile(x))$ \cite{raghu2019algorithmic} and use it with threshold. We also set up a random "rejector" as baseline.

\textbf{Cost versus Accuracy and Coverage.} In this experiment, we simulate the case where asking the server for help is costly regardless of whether the server's answer is correct or not.  We keep $c_1=1$ and examine the training of $\reject(x)$ and $\mobile(x)$ with different values of $c_e$. The result is shown in Figure \ref{fig:differentcost}. The x-axis means the current $c_e$ for different points. The three dotted lines refer to the accuracy if we only use $\edge(x)$, jointly use $\mobile(x)$, $\reject(x)$ and $\edge(x)$ or only use $\mobile(x)$. The solid line in purple refers to the coverage rate (portion of data that sent to server) evolves along with the cost $c_e$. The results show that Learning to Help can boost the performance by building collaboration between $\mobile(x)$ and $\edge(x)$. 

\textbf{Comparison with Confidence-based Methods.}\label{experiment:compare} By assigning different $c_e$, we collect the accuracy and coverage for the empirical minimizer for a fixed $\mobile(x)$, $\reject(x)$ case and a fixed $\mobile(x)$. For confidence-based methods and a random rejector, we collect the accuracy by setting up different thresholds for asking help. The comparison for accuracy over coverage rate is shown in Figure \ref{fig:compare}. From the results, we can find that learning to Help methods outperform confidence-based methods and jointly training $\reject(x)$ and $\edge(x)$ with fixed $\mobile(x)$ can mostly reach higher accuracy within the same training epochs compared to the result when we only train $\reject(x)$ with fixed $\mobile(x)$ and $\edge(x)$. 

In further experiments, shown in the supplementary Appendix ~\ref{appendix:experiment1}, we observe additional empirical training phenomena that coincide with our theoretical analysis.

\section{Conclusion}

In this work we proposed a Learning to Help framework which can be used as a ``version patch'' to prolong the functionality of legacy devices using complex ML models. 
This learning problem is not unique to mobile applications. Devices used in so-called ``smart infrastructures'' or industrial monitoring will also become restricted by the hardware limit or access authority~\cite{tennenhouse2023surprise} and may have to offload some computation.

We demonstrate the optimal rules and empirical implementation algorithm for Learning to Help. Our theories and experiments demonstrate the benefits of explicitly considering the trade-off between inference accuracy and other constraints (like latency). For future work, we aim to extend the one server case to multi-servers system where the machine learning models on different servers may work differently on certain subset of instances, motivated by the truth that human experts have expertise on different fields. 

\newpage
\section*{Acknowledgment}
The work of the authors was supported in part by the US National Science Foundation under award CNS-2148104.


\pdfoutput=1

\newpage

\onecolumn

\appendix
\subsection{Key assumptions for learning to help framework}\label{appendix:assumption}
 Here are the key assumptions for building our model:
\begin{enumerate}
    \item classifier $r$ and $e$ are jointly training while $m$ is pre-trained and fixed during the training process.
    \item we know the possibility that, on client classifier, the label of a given observation is positive: $q(x)=P(\mobile(x)=1 \mid X=x)$. Regression function $\eta(x)=P(Y=1|X=x)$ is draw from hidden distribution and isn't accessible to the trainer. 
    \item From probability and statistics perspective, the sample features, the label of sample and the output of three decision functions are random variables $X, Y, R, M, E$. We assume that some of they follow this Markov Chain: $Y\rightarrow X\rightarrow M$. That is, $P(M\neq Y|X=x)=P(M\neq Y|X=x,Y=y)$, or namely $P(M,Y|X=x)=P(M|X=x)P(Y|X=x)$, which means given $X=x$, $M$and $Y$ are independent. Then we can derive that:
    \begin{align}
        P(M\neq Y \mid X=x)&=1-P(M=Y|X=x)\nonumber\\
        &=q+\eta-2q\eta.\label{equ:P=q+eta-2qeta}
    \end{align}
\end{enumerate}
The proof detail for (\ref{equ:P=q+eta-2qeta}) is attached in appendix \ref{proof: accurate rate for client}. Since $\mobile(x)$ is pre-trained and fixed, $\mobile(x)$ only depends on the input sample $x$. Our assumption 2 for $q$ and assumption 3 that the classifier is independent to label also make sense in practice. Assumption 3 is essential for proof of the consistency of surrogate loss function $L_{\text{S}}$(defined by (\ref{equ:L2S})).

\subsection{Proof for Equation of $ P(M\neq Y \mid X=x)$}\label{proof: accurate rate for client}
    \begin{align}
        &P(M=Y \mid X=x)\nonumber \\ 
        &=P(M=1, Y=1| X=x)\nonumber\\
        &+P\left(M=0, Y={0} | X=x\right)\nonumber\\
        &=P(M=1|X=x)P(Y=1|X=x)\nonumber\\
        &+P(M=0|X=x)P(Y=0|X=x)\nonumber\\
        &=q\eta+(1-q)(1-\eta)\nonumber\\
        &=1-q-\eta+2q\eta.
    \end{align}
Then $P(M\neq Y \mid X=x)=q+\eta-2q\eta$.
\subsection{Proof for Bayes-optimal decision rules}\label{proof: bayes rule}
The classifier on edge is just a binary classifier without reject option, so the Bayes Rule for $\edge(x)$ is directly
\begin{align}
    e^{B}=\eta(x)-\frac{1}{2}.
\end{align}
For $r^{B}(x)$, we can derive it by comparing the posterior cost and plug in $e^{B}$:
\begin{align}
    0\cdot P(M=Y|X=x)+1\cdot P(M\neq Y|X=x) &\gtrless C_{1}(1-\eta) 1_{e^{B}>0} + C_{1}\eta  1_{e^{B}\le 0}+C_{e}\\
     P(M\neq Y|X=x)&\gtrless C_{1}\min{(\eta, 1-\eta)}+C_{e}\\
     P(M\neq Y|X=x)&\gtrless C_{1}\min{(\eta-\frac{1}{2}, \frac{1}{2}-\eta)}+C_{e}+\frac{C_{1}}{2}\\
     P(M\neq Y|X=x)-C_{e}-\frac{C_{1}}{2}&\gtrless -C_{1}|\eta -\frac{1}{2}|\\
     P(M\neq Y|X=x)-C_{e}-\frac{C_{1}}{2}+C_{1}|\eta -\frac{1}{2}|&\gtrless 0
\end{align}
When left side is greater than right side, reject. Otherwise, make decision locally.
Therefore, the reject function is:
\begin{align}
    r^{B}=C_{1}(\frac{1}{2}-|\eta-\frac{1}{2}|)-P(M\neq Y|X=x)+C_{e}
\end{align}

\subsection{Proof for Rademacher Upper bound}\label{proof: Rademacher natural}
Extending the result of theorem 3.3. in this book\cite{mohri2018foundations}, we got that
\begin{theorem}
Let $\mathcal{G}$ be a family of functions mapping from $\mathcal{Z}$ to $[a,b]$. Then, for any $\delta>0$, with probability at least $1-\delta$ over the draw of an i.i.d. sample $S$ of size $m$, each of the following holds for all $g \in \mathcal{G}$ :
$$
\begin{aligned}
\mathbb{E}[g(z)] & \leq \frac{1}{m} \sum_{i=1}^m g\left(z_i\right)+2 \Re_m(\mathcal{G})+\sqrt{\frac{(b-a)^{2}\log \frac{1}{\delta}}{2 m}} \\
\text { and } \quad \mathbb{E}[g(z)] & \leq \frac{1}{m} \sum_{i=1}^m g\left(z_i\right)+2 \widehat{\Re}_S(\mathcal{G})+3 \sqrt{\frac{(b-a)^{2}\log \frac{2}{\delta}}{2 m}}
\end{aligned}
$$
\end{theorem}
Since the value of natural loss is in $[0, c_1+c_{e}]$, we have with probability at least $1-\delta$, we have 
\begin{align}
    R\le \hat{R}+2\hat{\Re}_{n}(\mathcal{R}\times\mathcal{E})+3\sqrt{\frac{(c_{1}+c_{e})^{2}\ln{\frac{2}{\delta}}}{2n}}.
\end{align}
We should further pay attention to this term $\hat{\Re}_{n}(\mathcal{R}\times\mathcal{E})$:
\begin{align}
    \hat{\Re}_{n}(\mathcal{R}\times\mathcal{E})
    &=\text{E}_{\sigma}[\sup_{\mathcal{R, E}}\frac{1}{n}\sum_{i=1}^{n}\sigma_{i}(1_{m(x_{i}) y_{i} \leqslant 0} 1_{r(x_{i})>0}+c_1 1_{e(x_{i})\neq y_{i}} 1_{r(x_{i}) \leqslant0}+c_e 1_{r(x_{i}) \leqslant0})]\\
    &\le \text{E}_{\sigma}[\sup_{\mathcal{R}}\frac{1}{n}\sum_{i=1}^{n}\sigma_{i}(1_{m(x_{i}) y_{i} \leqslant 0} 1_{r(x_{i})>0}]\nonumber\\
    &+c_{1}\text{E}_{\sigma}[\sup_{\mathcal{R, E}}\frac{1}{n}\sum_{i=1}^{n}\sigma_{i}( 1_{e(x_{i})\neq y_{i}} 1_{r(x_{i}) \leqslant0})]\nonumber\\
    &+c_{e}\text{E}_{\sigma}[\sup_{\mathcal{R, E}}\frac{1}{n}\sum_{i=1}^{n}\sigma_{i}( 1_{r(x_{i}) \leqslant0})]\\
    &\le \text{E}_{\sigma}[\sup_{\mathcal{R}}\frac{1}{n}\sum_{i=1}^{n}\sigma_{i}1_{m(x_{i}) y_{i} \leqslant 0} ]+\text{E}_{\sigma}[\sup_{\mathcal{R}}\frac{1}{n}\sum_{i=1}^{n}\sigma_{i} 1_{r(x_{i})>0}]\nonumber\\
    &+c_{1}\text{E}_{\sigma}[\sup_{\mathcal{R, E}}\frac{1}{n}\sum_{i=1}^{n}\sigma_{i}( 1_{e(x_{i})\neq y_{i}} )]+c_{1}\text{E}_{\sigma}[\sup_{\mathcal{R, E}}\frac{1}{n}\sum_{i=1}^{n}\sigma_{i}(  1_{r(x_{i}) \leqslant0})]\nonumber\\
    &+c_{e}\text{E}_{\sigma}[\sup_{\mathcal{R, E}}\frac{1}{n}\sum_{i=1}^{n}\sigma_{i}( 1_{r(x_{i}) \leqslant0})]\\
    &\le\text{E}_{\sigma}[\sup_{\mathcal{R}}\frac{1}{n}\sum_{i=1}^{n}\sigma_{i}1_{m(x_{i}) y_{i} \leqslant 0} ]+\text{E}_{\sigma}[\sup_{\mathcal{R}}\frac{1}{n}\sum_{i=1}^{n}\sigma_{i} 1_{r(x_{i})>0}]\nonumber\\ 
    &+\frac{c_{1}}{2}\Hat{\mathfrak{R}}_{S}(\mathcal{E})\nonumber\\
    &+(c_{e}+c_{1})\text{E}_{\sigma}[\sup_{\mathcal{R, E}}\frac{1}{n}\sum_{i=1}^{n}\sigma_{i}( 1_{r(x_{i}) \leqslant0})]\\
    &\le 0+\frac{c_{e}+c_{1}+1}{2}\Hat{\mathfrak{R}}_{S}(\mathcal{R})+\frac{c_{1}}{2}\Hat{\mathfrak{R}}_{S}(\mathcal{E})
\end{align}
In sum, with probability at least $1-\delta$, 
\begin{align}
    R_{3}(r, e)\le \hat{R}_{3}(r, e)+(c_{e}+c_{1}+1)\Hat{\mathfrak{R}}_{S}(\mathcal{R})+c_{1}\Hat{\mathfrak{R}}_{S}(\mathcal{E})+3\sqrt{\frac{(c_{1}+c_{e})^{2}\ln{\frac{2}{\delta}}}{2n}}.
\end{align}
According to Hoeffding's inequality, with probability at least $1-\frac{\delta}{2}$, we have 
\begin{align}
    \hat{R}(r^{B}, e^{B})\le R(r^{B}, e^{B})+\sqrt{\frac{(c_{1}+c_{e})^{2}\ln{\frac{2}{\delta}}}{2n}}.
\end{align}
Let $\hat{r}^{*}, \hat{e}^{*}$ be the empirical minimizers, then we have:
\begin{align}
    \hat{R}(\hat{r}^{*}, \hat{e}^{*})\le\hat{R}(r^{B}, e^{B}).
\end{align}
Then the union bound is that, with probability at least $1-\frac{3}{2}\delta$, we have 
\begin{align}
    R(\hat{r}^{*}, \hat{e}^{*})\le R(r^{B},e^{B})+ (c_{e}+c_{1}+1)\Hat{\mathfrak{R}}_{S}(\mathcal{R})+c_{1}\Hat{\mathfrak{R}}_{S}(\mathcal{E})+4\sqrt{\frac{(c_{1}+c_{e})^{2}\ln{\frac{2}{\delta}}}{2n}}.
\end{align}
Here we didn't derive the inequalities with respect to the expected Rademacher Complexity. The reason is that in reality, we don't the true distribution of $(X, Y)$ so that we can't really calculate the value of the expected Rademacher Complexity.
\subsection{Proof for the derivation of surrogate loss function}\label{proof: Surrogate loss}
\begin{align}
    &L(r, m, e, x, y) \nonumber\\
    & =1_{\mobile(x) y \leqslant 0} 1_{\reject(x)>0}+(c_e+c_1) 1_{e\neq y} 1_{r \leqslant0}+c_e 1_{e= y} 1_{r \leqslant0}
    \nonumber\\ &=1_{\mobile(x) y \leqslant 0} 1_{\reject(x)>0}+c_1 1_{e\neq y} 1_{r \leqslant0}+c_e 1_{r \leqslant0}
    \nonumber\\ &=1_{\mobile(x) y \leqslant 0}  1_{-\reject(x)<0}+c_1 1_{e y \leqslant 0} 1_{r \leqslant 0}+c_e 1_{r \leqslant0}
    \nonumber\\ &\leqslant \max \left(1_{my \leqslant 0}1_{r\ge 0},c_1 1_{\max (e y, r) \leqslant 0}\right)+c_e 1_{r \leqslant0}
    \nonumber\\ &\leqslant \max \left(1_{my \leqslant 0}1_{-r\leqslant 0}, c_1 1 _{\frac{e y+r}{2} \leqslant 0}\right)+c_e 1_{r \leqslant0}
    \nonumber\\ &\leqslant \max \left(1_{my \leqslant 0}1_{-\alpha r\leqslant 0},c_1 1 _{\beta \frac{e y+r}{2} \leqslant 0}\right)+c_e 1_{r \leqslant0}
    \nonumber\\ &\leqslant \max \left(1_{my \leqslant 0}\phi (\alpha r),c_1 \psi(-\beta \frac{e y+r}{2} )\right)+c_e 1_{r \leqslant0}
    \nonumber\\ &\leqslant \mathbf{1}_{my \le 0}\phi( \alpha r )+c_1 \psi(-\beta \frac{e y+r}{2})+c_e \exp{(-r)}
\end{align}
where $\phi(\cdot)$ and $\psi(\cdot)$ can be any functions that satisfy $\mathbf{1}_{z}\le \phi(z)$ and $\mathbf{1}_{z}\le \psi(z)$. We choose exponential function as \(\phi\) and\(\psi\) then the surrogate loss is:
\begin{equation}
L_{\text{S}}=1_{my \le 0}\exp \left(\alpha r\right)+c_1 \exp \left(\frac{\beta}{2}(-e y-r)\right)+c_e \exp({-r})
\end{equation}

\subsection{Proof of the calibration of surrogate loss}\label{Proof:Calibration}

The surrogate loss in this case is
\begin{equation}
L_{\text{S}}=1_{my \leqslant 0}\exp \left(\alpha r\right)+c_1 \exp \left(\frac{\beta}{2}(-e y-r)\right)+c_e e^{-r}
\end{equation}
where $\alpha>0$,$\beta>0$. As \(h\) is fixed classifier that we won't change  during the training, we can regard it as a random variable. Its conditional distribution is \(q(x)=P(M=1|X=x)\) and we define \(\eta (x)=P(Y=1|X=x)\).
\\Then we need to find the infimum of expected loss over space \(\mathcal{R} \times \mathcal{M}\):
\begin{equation}
\begin{aligned} \inf _{e, r} \mathbb{E}_{x, y, m}\left[L_{\text{S}}(m, r, x, y, e)\right] &=\inf _{e, r} \mathbb{E}_{x} \mathbb{E}_{y \mid x} \mathbb{E}_{m \mid x, y}\left[L_{\text{S}}(m, r, x, y, e)\right] \\ &=\mathbb{E}_{x} \inf _{\edge(x), \reject(x)} \mathbb{E}_{y \mid x} \mathbb{E}_{m \mid x, y}\left[L_{\text{S}}(\mobile(x), \reject(x), x, y, \edge(x))\right] \end{aligned}
\end{equation}

For each given x, the expected loss is:
\begin{equation}
\begin{array}{l}E_{y | x} E_{m |x, y}[L _{\text{S}} ] 
\\ =\eta(x) q(x)\left[0+C_{1}\exp \left(\frac{\beta}{2}(-e-r)\right)+C_{e}\exp{(-r)}\right]
\\+ (1-\eta(x)) q(x)\left[\exp \left(\alpha r\right)+C_{1}\exp \left(\frac{\beta}{2}(e-r)\right)+C_{e}\exp{(-r)}\right]
\\+(1-\eta(x)) (1-q(x))\left[0+C_{1}\exp \left(\frac{\beta}{2}(e-r)\right)+C_{e}\exp{(-r)}\right]
\\+\eta(x) (1-q(x))\left[\exp \left(\alpha r\right)+C_{1}\exp \left(\frac{\beta}{2}(-e-r)\right)+C_{e}\exp{(-r)}\right]
\end{array}\label{expectedloss3}
\end{equation} 
Bayes Rule for this setting is:
\begin{equation}
\begin{array}{l}
e^{B}(x)=\eta(x)-\frac{1}{2}
\end{array}
\end{equation}
\begin{equation}
\begin{array}{l}
r^{B}(x)=C_1(\frac{1}{2}-|\eta(x)-\frac{1}{2}|)-P(M\neq Y|X=x)+C_{e} \label{equ:rBC_3}
\end{array}
\end{equation}
Our target is to match \(e^{B}(x), r^{B}(x)\) with \(e^{*}(x), r^{*}(x)\), namely, \(sign(e^{B}(x))=sign(e^{*}(x))\), \(sign(r^{B}(x))=sign(r^{*}(x))\), for each given $x$, by choosing specific $\alpha$ and $\beta$.
\\
From the construction of \(L_{PH}\), we know that \(L_{PH}\) is convex for \(e\) and \(r\) and it's differentiable. We can take the partial derivatives for them, respectively. First, we take the partial derivative over \(e\):

\begin{align}
    \frac{\partial E_{y | x} E_{e |x, y}[L _{PH} ]}{\partial e}&=0,    
\end{align}
which is,
\begin{align*}
    -C_{1}\frac{\beta}{2} \eta \exp \left(\frac{\beta}{2}(-e-r)\right)+C_{1}\frac{\beta}{2}(1-\eta) \exp \left(\frac{\beta}{2}(e-r)\right)&=0 
    \\ \eta \exp \left(\frac{-\beta}{2}(e+r)\right)-(1-\eta) \exp \left(\frac{\beta}{2}(e-r)\right)&=0 
    \\ \ln \eta-\frac{\beta}{2}(e+v)-\ln (1-\eta)+\frac{\beta}{2}(e-r)&=0.
\end{align*}
\begin{align}
    \Rightarrow e^{*}&=\frac{\ln(\frac{\eta}{1-\eta})}{\beta}.
\end{align}
Apparently, we have $\text{sign}(e^{B})=\text{sign}(e^{*})$ always hold for any $\eta(x)\in (0, 1)$ and $\beta>0$. We define $\ln{0}=-\infty$. Then for the extreme cases: if $\eta(x)=1$, we have $\text{sign}(e^{*})=\text{sign}((\ln{1}-\ln{0})/\beta)=1=\text{sign}(e^{B})$ and if $\eta(x)=1$, we have $\text{sign}(e^{*})=\text{sign}(\ln{0}/\beta)=-1=\text{sign}(e^{B})$. Therefore, $\text{sign}(e^{*})$ is consistent for any $\eta(x)\in [0, 1]$ and $\beta>0$ now.\\
Then we take partial derivative over $r$:
\begin{align}
    \frac{\partial E_{y | x} E_{m |x, y}[L _{PH}]}{\partial r}&=0.
\end{align}
After simplifying, the equation becomes
\begin{align}
    C_{1}\eta \cdot\left(-\frac{\beta}{2}\right) \exp \left(-\frac{\beta}{2}(e+r)\right)+(q-\eta q+\eta-\eta q) \alpha e^{\alpha r}+(1-\eta)C_{1}\left(-\frac{\beta}{2}\right) \exp \left(-\frac{\beta}{2}(r-e)\right)&=C_{e}\exp{(-r)}.
\end{align}
Plug in \(e^*\), then we get
\begin{align}
&-\frac{\beta}{2}C_{1} \eta \exp \left(-\frac{\beta}{2} r\right) \exp (-\frac{\beta}{2} \cdot \frac{\ln \left(\frac{\eta}{1-\eta}\right)}{\beta})\nonumber\\
&+(q+\eta-2 \eta q) \alpha e^{\alpha r} -C_{1}\frac{\beta}{2}(1-\eta) \exp \left(-\frac{\beta}{2} r\right) \exp \left(\frac{\beta}{2} \frac{\ln \left(\frac{\eta}{1-\eta}\right)}{\beta}\right)=C_{e}\exp{(-r)}. \label{equ:rstar_C1Ce}
\end{align}

\begin{align}
    \Rightarrow(q+\eta-2 \eta q) \alpha e^{\alpha r}=C_{1}\beta e^{-\frac{\beta}{2} r} \cdot \sqrt{\eta(1-\eta)}+C_{e}\exp{(-r)}\label{equ:r_pluginmstar}
\end{align}
We also assume \(P(M\neq Y|X=x)=P(M\neq Y|X=x,Y=y)\), namely \(P(M,Y|X=x)=P(M|X=x)P(Y|X=x)\). Then $P(M\neq Y| X=x)=q+\eta-2q\eta$. To simplify the calculation, we let $\beta=2$. Plug in \ref{equ:r_pluginmstar} then we get:
\begin{align}
    r^{*}=\frac{1}{\alpha+1} \ln \left(\frac{2C_{1} \sqrt{\eta(1-\eta)}+C_{e}}{\alpha P(M \neq Y \mid X=x).}\right)
\end{align}
Now the case is that we know $C_1$, $C_e$ and $P(M \neq Y \mid X=x)$ and we should choose $\alpha$ to make $r^{*}$ consistent with $r^{B}$ no matter what $\eta(x)$ is. Since $P(M \neq Y \mid X=x)\le C_e$ varies with different $x$, let's divide this problem into three cases where $P(M \neq Y \mid X=x)\le C_e$, $P(M \neq Y \mid X=x)> C_e+\frac{1}{2}C_1$ and $C_e<P(M \neq Y \mid X=x)\le C_e+\frac{1}{2}C_1$.
\subsubsection{When $P(M \neq Y \mid X=x)\le C_e$}
Since $P(M \neq Y \mid X=x)\le C_e$, we have $r^{B}\ge0$ always hold. Then we need find $\alpha$ to ensure $r^{*}\ge 0$ as well.
\begin{align}
    &r^{*}=\frac{1}{\alpha+1} \ln \left(\frac{2C_{1} \sqrt{\eta(1-\eta)}+C_{e}}{\alpha P(M \neq Y \mid X=x)}\right)\ge0\\
    &\Leftrightarrow 2C_{1} \sqrt{\eta(1-\eta)}+C_{e}\ge \alpha P(M \neq Y \mid X=x)\\
    &\Leftarrow C_e\ge \alpha P(M \neq Y \mid X=x)
\end{align}
Since we already have $P(M \neq Y \mid X=x)\le C_e$, just let $\alpha \le 1$ can ensure $r^{*}\ge0$.

\subsubsection{When $P(M \neq Y \mid X=x)> C_e+\frac{1}{2}C_1$}
When $P(M \neq Y \mid X=x)> C_e+\frac{1}{2}C_1$, we have $r^{B}\le0$ always hold. Then we need find $\alpha$ to ensure $r^{*}\le 0$ as well.
\begin{align}
    &r^{*}=\frac{1}{\alpha+1} \ln \left(\frac{2C_{1} \sqrt{\eta(1-\eta)}+C_{e}}{\alpha P(M \neq Y \mid X=x)}\right)\le0\\
    &\Leftrightarrow 2C_{1} \sqrt{\eta(1-\eta)}+C_{e}\le \alpha P(M \neq Y \mid X=x)\\
    &\Leftarrow \frac{1}{2}C_1+C_e\le \alpha P(M \neq Y \mid X=x)
\end{align}
Since we already have $P(M \neq Y \mid X=x)> C_e+\frac{1}{2}C_1$, let $\alpha>1$ is enough to ensure $r^{*}\le0$.

\subsubsection{$C_e<P(M \neq Y \mid X=x)\le C_e+\frac{1}{2}C_1$}
This case is more complicated because $r^{B}$ could either be non-negative or negative according to the value of $\eta(x)$ which we don't know. Then let's do the following analysis:\\
According to the formula of \(r^{B}\), we get \(\eta(1-\eta)=\frac{1}{4}-(\frac{1}{2}-\frac{r^{B}+P(M \neq Y \mid X=x)-C_{e}}{C_1})^2\) and \(C_{e}-P(M \neq Y \mid X=x)\le r^{B}\le \frac{C_{1}}{2}+C_{e}-P(M \neq Y \mid X=x)\), since $\eta\in [0,1]$.
\\If \(r^{B}\ge 0\), we need \(r^{*}\ge 0\), which is 
\begin{align}
    r^{B}\ge 0 &\Rightarrow \frac{2C_1\sqrt{\eta(1-\eta)}+C_{e}}{\alpha} \ge P(M \neq Y \mid X=x)
    \\&\Rightarrow \frac{1}{\alpha}\ge\frac{P(M \neq Y \mid X=x)}{2C_1\sqrt{\eta(1-\eta)}+C_{e}}
    \\&\Rightarrow \frac{1}{\alpha}\ge \frac{P(M \neq Y| X=x)}{2C_1\sqrt{\frac{1}{4}-(\frac{1}{2}-\frac{r^{B}+P(M \neq Y \mid X=x)-C_{e}}{C_1})^2}+C_{e}}
    \\&\Rightarrow \frac{1}{\alpha}\ge \frac{P(M \neq Y| X=x)}{\sqrt{C_{1}^{2}-(C_{1}+2C_{e}-2r^{B}-2P(M \neq Y \mid X=x))^2}+C_{e}}
    \\&\Leftrightarrow \frac{\sqrt{C_{1}^{2}-(C_{1}+2C_{e}-2r^{B}-2P(M \neq Y \mid X=x))^2}+C_{e}}{P(M \neq Y| X=x)}\ge \alpha \label{eq:rstargreaterthan0}
\end{align}
We define \(f(r^{B})=\frac{\sqrt{C_{1}^{2}-(C_{1}+2C_{e}-2r^{B}-2P(M \neq Y \mid X=x))^2}+C_{e}}{P(M \neq Y| X=x)}\) with $r^{B}\in [-P+C_{e},C_1/2-P+C_{e}]$ (according to the formula of $r^{B}$). The shape of $f(r^{B})$ can be form in three possible sub-cases, depending on the value of constants:
    \begin{figure}[bthp]
        \centering
        \includegraphics[width=10cm]{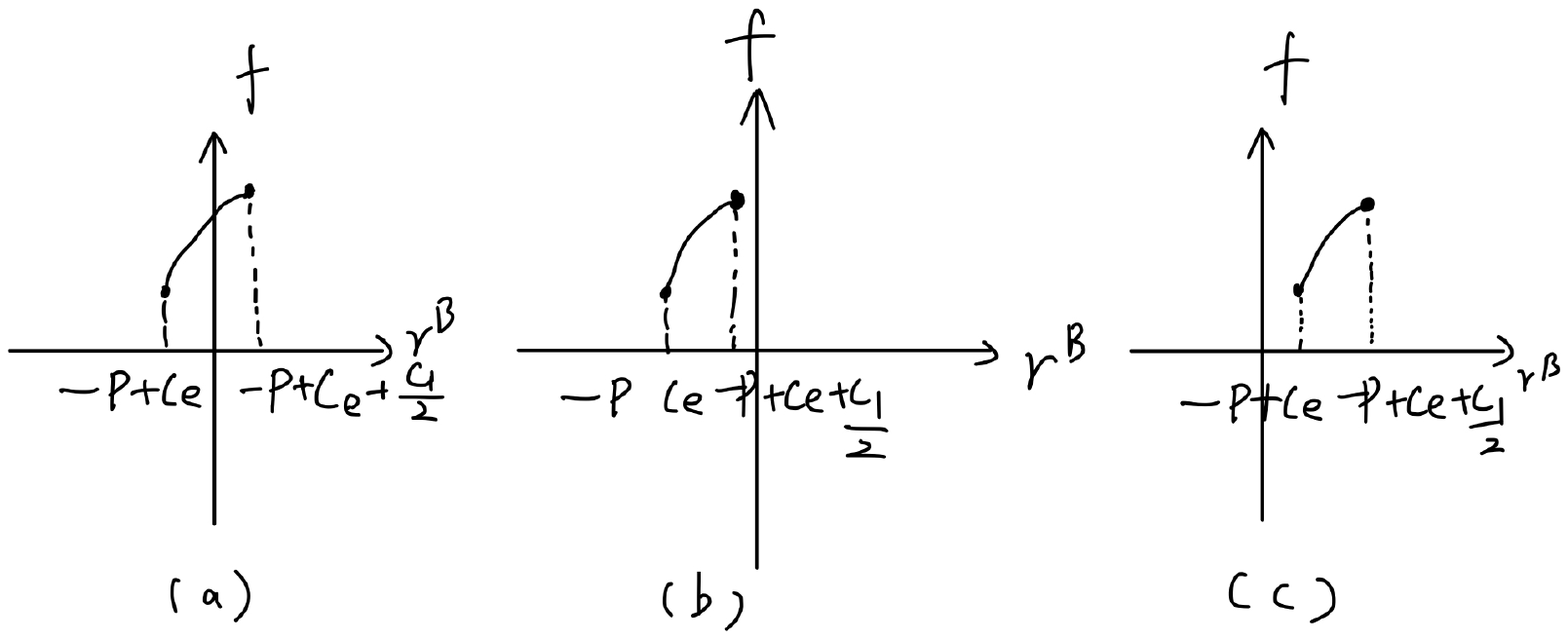}
    \end{figure}
Here, $C_e<P(M \neq Y \mid X=x)$ means $-P(M \neq Y \mid X=x)+C_e<0$ so sub-case (c) is excluded. Since we consider $r^{B}\ge0$ at this moment, sub-case (b) is excluded as well and for sub-case (a), we only consider this domain $r^{B}\in[0, C_1/2-P+C_{e}]$. Then $\min f(r^{B})=f(0)$. To make the inequality (\ref{eq:rstargreaterthan0}) always holds, we can let $\min f(r^{B})\ge \alpha$. 
\begin{align}
    \min f(r^{B})&\ge \alpha\\
    f(0)&\ge \alpha \\
    \alpha&\le\frac{c_e+\sqrt{4c_1 P-4c_1 c_e-4P^2+8c_e P-4C_{e}^{2}}}{P}
\end{align}
\\If \(r^{B}\le 0\), we need \(r^{*}\le 0\). Similarly, we require that 
\begin{align}
\frac{\sqrt{C_{1}^{2}-(C_{1}+2C_{e}-2r^{B}-2P(M \neq Y \mid X=x))^2}+C_{e}}{P(M \neq Y| X=x)}\le \alpha \label{eq:rstarsmallerthan0}
\end{align}
Here, $P(M \neq Y \mid X=x)\le C_e+\frac{1}{2}C_1$ mean $-P(M \neq Y \mid X=x)+C_e+\frac{1}{2}C_1\ge0$ so sub-case (b) is excluded. Since we consider $r^{B}\le0$ , sub-case (c) is excluded and for sub-case (a), we only consider this domain $r^{B}\in[-P+C_{e}, 0]$. Then $\max f(r^{B})=f(0)$. To make the inequality (\ref{eq:rstarsmallerthan0}) always holds, we can let $\max f(r^{B})\le \alpha$, that is
\begin{align}
    \max f(r^{B})&\le \alpha\\
    f(0)&\le \alpha \\
    \alpha&\ge\frac{c_e+\sqrt{4c_1 P-4c_1 c_e-4P^2+8c_e P-4C_{e}^{2}}}{P}.
\end{align}

In order to ensure '$r^{B}\le 0 \Rightarrow r^{*}\le0$' and '$r^{B}\ge 0 \Rightarrow r^{*}\ge0$' simultaneously, the only available setting is $\alpha=\frac{c_e+\sqrt{4c_1 P-4c_1 c_e-4P^2+8c_e P-4C_{e}^{2}}}{P}$, then the consistency of the surrogate loss is ensured.\\

In sum, during training processing, for any instance $x$, 
\begin{itemize}
    \item if $P(M \neq Y \mid X=x)\le C_e$, let $\alpha \le 1$;
    \item if $P(M \neq Y \mid X=x)> C_e+\frac{1}{2}C_1$, let $\alpha >1$;
    \item if $C_e<P(M \neq Y \mid X=x)\le C_e+\frac{1}{2}C_1$, let $\alpha=\frac{c_e+\sqrt{4c_1 P-4c_1 c_e-4P^2+8c_e P-4C_{e}^{2}}}{P}$;
\end{itemize}
then we can ensure the optimizer of this surrogate loss function is consistent with Bayes Classifier.
\subsection{The empirical algorithm for surrogate loss function}
\subsubsection{Training Stage}
The theorem \ref{surrogateLoss3:calibration} gives us a guideline to train our rejector $r(x)$ and server classifier $e(x)$. However, in practice, we don't have the knowledge of $P(x)$ which relies on the unknown distribution $\mathbf{D}$. So we propose this empirical algorithm for the training stage.
\begin{algorithm}\label{algo: empirical}

\caption{Empirical Algorithm For Our Surrogate Loss Function}
\begin{algorithmic}
\REQUIRE Training Set $\left\{\left(x_i, y_i): i \in[n]\right\}\right.$, Fixed Client Classifier $m$, Rejector $r^{0}$, Sever Classifier $e^{0}$, Hyper-parameters: $c_e$,  $c_1$, $0\le \alpha_1\le1$, $\alpha_{2}>1$.
\FOR{$i = 1$ to $n$}
\STATE $\Hat{P}(x_i)\leftarrow\text{\textbf{EstimatePx}}(m,x_i,y_i)$
\STATE $\beta\leftarrow2$
\IF{$\Hat{P}(x_i)\le c_e$}
\STATE $\alpha\leftarrow\alpha_1$
\ELSIF{$\Hat{P}(x_i)> c_e+\frac{1}{2}c_1$}
\STATE $\alpha\leftarrow\alpha_2$
\ELSE
\STATE $\alpha\leftarrow\alpha=\frac{c_e+\sqrt{4c_1 \Hat{P}(x_i)-4c_1 c_e-4\Hat{P}(x_i)^2+8c_e \Hat{P}(x_i)-4C_{e}^{2}}}{\Hat{P}(x_i)}$
\ENDIF
\STATE $L_{\text{S}}^{i}\leftarrow c_1 \exp \left(\frac{\beta}{2}(-e^{i-1}(x_{i}) y_{i}-r^{i-1}(x_{i})\right)+c_e \exp({-r^{i-1}(x_{i})})+1_{m(x)y_{i} \le 0}\exp \left(\alpha r^{i-1}(x_{i})\right)$
\STATE $r^{i}, e^{i}\leftarrow \textbf{Optimizer}(L_{\text{S}}^{i}, r^{i-1}, e^{i-1})$
\ENDFOR
\RETURN $r^{n}, e^{n}$
\end{algorithmic}
\end{algorithm}
In this algorithm, \textbf{EstimatePx} can be any function that estimates the probability $P(M\neq Y|X=x)$. In our experiment, we add a Sigmoid function to the output of $m$ and the give the estimate, depending on the true label $y$. \textbf{Optimizer} can be any optimization methods that fit for convex and differentiable loss function. Here in experiments we use Stochastic gradient descent(SGD).
\begin{algorithm}\label{algo: estimatePx}
\caption{\textbf{EstimatePx} Estimation Method for $P(M \neq Y \mid X=x)$ in Empirical Experiments}
\begin{algorithmic}
\REQUIRE Fixed Client Classifier $m$, current training tuple $x_i$ and $y_i$.
\IF{$y_i==1$}
\STATE $\Hat{P}(x_i)\leftarrow 1-\frac{1}{1+\exp{(m(x_i))}}$
\ELSIF{$y_i==-1$}
\STATE $\Hat{P}(x_i)\leftarrow\frac{1}{1+\exp{(m(x_i))}} $
\ENDIF
\RETURN $\Hat{P}(x_i)$
\end{algorithmic}
\end{algorithm}

\subsubsection{Testing Stage}
Since we already got the rejector and server classifier, we don't need calculate the loss function and estimate $P(x)$ anymore. In testing stage, a new $x$ firstly goes to rejector, if $r\le0$, then we send it to server and the output is $e(x)$, otherwise ($r>0$), we just send it to client classifier and the output is $m(x)$. For the related application in mobile edge inference system, we can assume that $c_e$ is the latency cost. Then the inference decision rule is a trade-off between accuracy and latency since $y_{\text{client}}$ is less accurate but more instant while $y_{\text{server}}$ is more accurate but reacts slower. 
\begin{algorithm}\label{algo: inference}
\caption{Our Proposed Method for Prediction on New Instance $x$}
\begin{algorithmic}
\REQUIRE Fixed Client Classifier $m$, Trained Rejector $r^{n}$, Trained Server Classifier $e^{n}$, input instance $x$.
\IF{$r^{n}(x)\le 0$}
\STATE $y_{\text{server}\leftarrow e^{n}(x)}$
\STATE Prediction$\leftarrow y_{\text{server}}$
\ELSIF{$r^{n}(x)> 0$}
\STATE $y_{\text{client}\leftarrow m(x)}$
\STATE Prediction$\leftarrow y_{\text{client}}$
\ENDIF
\RETURN Prediction
\end{algorithmic}
\end{algorithm}

\subsection{Complementary Experiment Result: Empirical Training Phenomena}\label{appendix:experiment1}
\begin{table*}[htbp]
 \caption{Cross check of the accuracy on different test set portions partitioned by rejector,unit percentage "\%"}
\begin{small}
  \centering
  \resizebox{\textwidth}{!}{
  \begin{tabular}{|c|c|c|c|c|c|c|c|c|c|c|c|c|c|c|c|c|c|c|}
    \hline
    \multirow{2}{*}{} & \multicolumn{3}{c|}{airplane vs car} & \multicolumn{3}{c|}{bird vs cat} & \multicolumn{3}{c|}{deer vs dog} & \multicolumn{3}{c|}{frog vs horse}& \multicolumn{3}{c|}{ship vs truck}\\
    \cline{2-16}& $m$ & $e$& differ. & $m$ & $e$&differ. & $m$ &  $e$&differ.&  $m$& $e$&differ.& $m$& $e$&differ. \\ \hline
    data not deferred  &87.9 &94.9&\textbf{+7.0} &76.3 &80.0&\textbf{+3.7} &78.5 &88.1&\textbf{+9.6 }&87.2 & 95.1&\textbf{+7.9} &87.4&95.1&\textbf{+7.7} \\ \hline
    data deferred  & 46.6&92.2&\textbf{+45.6} &43.0 &79.5&\textbf{+36.5} &35.4 &82.6&\textbf{+47.2 }&19.3 &94.5&\textbf{+75.2} &42.7&90.8&\textbf{+48.1}\\ \hline
  \end{tabular}
  }
 
  \label{table:why learn to help is good}
  \end{small}
\end{table*}
From the result in \ref{experiment:compare}, we are curious why jointly training is better separately training. We did following experiment for Learning to Help with fixed $\mobile(x)$ model. We analyze the data that sent to different classifier by $\reject(x)$. For those data that is supposed to be sent to server, we additionally test their accuracy on $\mobile(x)$. For those data that is supposed to stay at client, we additionally test their accuracy on $\edge(x)$. Then we get the table show in \ref{table:why learn to help is good}. The column differ. means the difference of accuracy on different classifier for the same portion of data. From the table, we find that the performance for those data that is kept locally is close while those data sent to server by $\reject(x)$ is extremely bad if predicted by $\mobile(x)$. In this sense, rejector $\reject(x)$ learns to pick up the ambiguous samples or outliers that are hard for $\mobile(x)$ to identify and $\edge(x)$ are fed with "harder" cases so it "grows" faster than separately trained model. Both $\reject(x)$ and $\edge(x)$ benefit from jointly training.

\end{document}